\documentclass[10pt,twocolumn,letterpaper]{article}

\usepackage[pagenumbers]{iccv} % To force page numbers, e.g. for an arXiv version
\usepackage{amsmath} % assumes amsmath package installed
\usepackage{graphicx}
\usepackage{caption}
\usepackage{longtable}
\usepackage{booktabs}
\usepackage{cuted}
\usepackage{capt-of}
\usepackage{adjustbox}
\usepackage{xcolor}
\usepackage{pifont} % for tick and cross symbols
\usepackage{multirow}
\usepackage{colortbl}
\usepackage{afterpage}
\usepackage{makecell} % For wrapping text in cells
\usepackage{array}
\usepackage{amssymb}

\newcommand{\up}{\color{green!60!black}\uparrow}
\newcommand{\down}{\color{red!60!black}\downarrow}

\definecolor{iccvblue}{rgb}{0.21,0.49,0.74}
\usepackage[pagebackref,breaklinks,colorlinks,allcolors=iccvblue]{hyperref}
\def\paperID{00000} % *** Enter the Paper ID here
\def\confName{ICCV}
\def\confYear{2025}

\title{ClaraVid: A Holistic Scene Reconstruction Benchmark From Aerial Perspective With Delentropy-Based Complexity Profiling}
\author{Radu Beche\\
Technical University of Cluj-Napoca\\
Cluj-Napoca, Romania\\
{\tt\small radu.beche@cs.utcluj.ro}
\and
Sergiu Nedevschi\\
Technical University of Cluj-Napoca\\
Cluj-Napoca, Romania\\
{\tt\small sergiu.nedevschi@cs.utcluj.ro}
}
\begin{document}
\newcolumntype{C}[1]{>{\centering\arraybackslash}p{#1}}
\maketitle

\begin{strip}\centering
\vspace{-0.8cm}
\includegraphics[width=\textwidth]{assets/001_overview_claravid.pdf}
\captionof{figure}[.]{\textbf{Overview of ClaraVid}:
    (a) \textit{Multi-viewpoint UAV acquisition:} High-resolution aerial imagery is captured from multiple altitudes, ensuring diverse perspectives. 
    (b) \textit{High-fidelity, diverse environments:} complex urban, suburban, and natural landscapes. 
    (c) \textit{Multimodal ground truth:} Pixel-level and scene-level multimodal data for holistic scene reconstruction and semantic mapping.
    \label{fig:claravid_overview}}
\label{fig:overview1}
\end{strip}
\begin{abstract}

The development of aerial holistic scene understanding algorithms is hindered by the scarcity of comprehensive datasets that enable both semantic and geometric reconstruction. While synthetic datasets offer an alternative, existing options exhibit task-specific limitations, unrealistic scene compositions, and rendering artifacts that compromise real-world applicability. We introduce ClaraVid, a synthetic aerial dataset specifically designed to overcome these limitations. Comprising 16,917 high-resolution images captured at 4032×3024 from multiple viewpoints across diverse landscapes, ClaraVid provides dense depth maps, panoptic segmentation, sparse point clouds, and dynamic object masks, while mitigating common rendering artifacts. To further advance neural reconstruction, we introduce the Delentropic Scene Profile (DSP), a novel complexity metric derived from differential entropy analysis, designed to quantitatively assess scene difficulty and inform reconstruction tasks. Utilizing DSP, we systematically benchmark neural reconstruction methods, uncovering a consistent, measurable correlation between scene complexity and reconstruction accuracy. Empirical results indicate that higher delentropy strongly correlates with increased reconstruction errors, validating DSP as a reliable complexity prior. The data and code are available on the project page: \href{https://rdbch.github.io/claravid/}{rdbch.github.com/claravid}.
\end{abstract}
    
\section{Introduction}

% problem context and motivation
Semantic and geometric scene understanding derived from low-altitude uncrewed aerial vehicle (UAV) imagery has become increasingly essential for applications such as disaster response ~\cite{zhu2021visdrone}, urban mapping \cite{nex2014uavmapping}, and infrastructure inspection \cite{rymer2020uavpowerline}, where accurate and high-fidelity environmental perception directly informs critical decision-making processes. Recent advancements in neural reconstruction techniques, notably Neural Radiance Fields (NeRFs)\cite{mildenhall2020nerf, ost2021nsg, kerr2023lerf} and Gaussian Splatting (GSplat) approaches\cite{kerbl2023gsplat,chen2023omnire,qin2023langsplat}, have demonstrated significant potential in enabling scene-level querying and novel-view synthesis, offering a transformative approach to reconstructing complex 3D environments from sparse 2D observations. However, evaluating such holistic reconstruction of urban environments—encompassing both semantic and geometric aspects—remains constrained by the limitations of existing datasets. Real-world benchmarks providing comprehensive multi-modal ground truth for the joint tasks are inexistent for urban scenarios, and synthetic alternatives suffer from critical shortcomings: they often lack key sensing modalities such as dense semantic or instance segmentation \cite{li2023matrixcity, liqiang2022UrbanScene3D}, are not collected for scene mapping applications \cite{rizzoli2023syndrone, khose2024skyscenes}, or are captured from restricted viewpoints that hinder a full assessment of reconstruction quality and generalization \cite{kolbeinsson2024ddos}. 

% claravid introduction
In this work, we introduce ClaraVid, a synthetic aerial dataset crafted to support the training and evaluation of holistic scene reconstruction while also enabling lower-level perception tasks. ClaraVid employs a mapping-centric approach, emulating UAV mapping missions to reflect real-world aerial survey workflows. Spanning 5 distinct environments across 8 missions, the dataset provides high-resolution imagery, collected simultaneously from 3 different viewpoints, accompanied by multimodal annotations: dense depth maps, panoptic segmentation and dynamic object masks to delineate moving entities--a distinctive enhancement over prior synthetic datasets. Furthermore, it includes multi-resolution multi-modal scene-level point clouds approximating LiDAR scans, broadening its utility for geometric analysis. Data quality is ensured through an optimized rendering pipeline that addresses challenges such as foliage culling and level-of-detail inconsistencies, resulting in enhanced fidelity for scene reconstruction and semantic segmentation. Notably, Claravid systematically spans structural complexities--from homogeneous areas to highly detailed urban environments--quantified via scene complexity profile. The dataset is complemented by a benchmarking framework designed to assess semantic and geometric reconstruction accuracy at various viewpoints, while also supporting frame-level tasks (depth estimation, panoptic segmentation, etc). An overview is presented in \autoref{fig:overview1}.

% dsp introduction
A major challenge in UAV-based neural scene reconstruction is quantifying scene complexity in a way that correlates with the difficulty of downstream reconstruction tasks (e.g., novel view synthesis) only from 2D measurements. Currently, the base solution relies on heuristic assessments rather than formalized metrics, making it difficult to systematically predict the impact of scene structure on the reconstruction. Delentropy\cite{larkin2016delentropy}, an entropy measure derived from gradient field analysis, has previously been employed to analyze information content in pixel-level tasks \cite{rahane2020delentropydataset, tunon2024subpipedelentropy, cagas2024medicaldelentropy}. In this work, we adapt delentropy to quantify scene complexity in neural reconstruction and  introduce the \emph{Delentropic Scene Profile} as a descriptor of scene-level complexity. We empirically demonstrate a strong correlation between the proposed metric and the final error across multiple reconstruction approaches. 

% contributions
To summarize, our paper introduces the following contributions to advance UAV-based holistic scene understanding: (1) We present ClaraVid, a novel synthetic aerial dataset specifically designed for holistic neural scene reconstruction, comprising high-resolution imagery, multi-modal annotations, and realistic rendering tailored for neural scene reconstruction tasks from an aerial perspective. (2) We introduce the Delentropic Scene Profile, a delentropy-based metric for quantitatively assessing scene complexity only from 2D measurements. (3) We empirically demonstrate the utility of both the dataset and DSP through extensive experiments.

% table that won't stay where I want it to stay so I have to put it here
\begin{table*}[t]
\centering
\caption{\textbf{Related Aerial Datasets}--Comparison of datasets from aerial and automotive domains relevant for holistic scene representation.}
\resizebox{\linewidth}{!}{%
\begin{tabular}{l c c >
{\centering\arraybackslash}p{1.2cm} >{\centering\arraybackslash}p{1.3cm} c >{\centering\arraybackslash}p{1.2cm} >{\centering\arraybackslash}p{1.5cm} >{\centering\arraybackslash}p{1.4cm} c >{\centering\arraybackslash}p{1.4cm} >{\centering\arraybackslash}p{1.4cm} c}
\toprule
\textbf{Dataset} 
& \textbf{Year} 
& \textbf{Type*} 
& \textbf{Camera Calib.}  
& \textbf{Seg.}  
& \textbf{Depth}  
& \textbf{Scene PCL}  
& \textbf{Dynamic Mask}  
& \textbf{For Mapping}  
& \textbf{Viewpoints}  
& \textbf{Render Issues}  
& \textbf{Image Size}  
& \textbf{Samples} \\
\midrule
\multicolumn{13}{l}{\textit{\textbf{Automotive Datasets}}} \\
% Kitti 360 \cite{liao2022kitti} & 2022 & R & \textcolor{green}{\ding{51}} & Panoptic & \textcolor{green}{\ding{51}} & - & - & - & 4 & - & 1408x376 & 320000 \\
A2D2 \cite{geyer2020a2d2} & 2020 & R & \textcolor{green}{\ding{51}} & Panoptic & \textcolor{green}{\ding{51}} & - & - & - & 6 & - & 1440x960 & 41.2k \\
Waymo \cite{sun2020scalability} & 2020 & R & \textcolor{green}{\ding{51}} & Panoptic & \textcolor{green}{\ding{51}} & - & - & - & 5 & - & 1920x1280 & 390k \\
% GTA5 \cite{richter2017gta5} & 2017 & S & \textcolor{red}{\ding{55}} & Panoptic & \textcolor{green}{\ding{51}} & - & - & - & 1 & Medium & 1980x1080 & 128k \\
% Synscapes \cite{wrenninge2018synscapes} & 2018 & S & \textcolor{red}{\ding{55}} & Panoptic & \textcolor{green}{\ding{51}} & - & - & - & 1 & Low & 2048x1024 & 25000 \\
Urban Syn \cite{gómez2023urbansyn} & 2023 & S & \textcolor{red}{\ding{55}} & Panoptic & \textcolor{green}{\ding{51}} & - & - & - & 1 & Low & 2048x1024 & 7.5k \\
\midrule
\multicolumn{13}{l}{\textit{\textbf{Aerial Datasets}}} \\
UDD \cite{chen2018udd} & 2018 & R & \textcolor{red}{\ding{55}} & Semantic & \textcolor{red}{\ding{55}} & \textcolor{red}{\ding{55}} & \textcolor{red}{\ding{55}} & \textcolor{red}{\ding{55}} & 1 & - & 4096x2160 & 0.3k \\
Aeroscapes \cite{nigam2018aeroscapes} & 2019 & R & \textcolor{red}{\ding{55}} & Semantic & \textcolor{red}{\ding{55}} & \textcolor{red}{\ding{55}} & \textcolor{red}{\ding{55}} & \textcolor{red}{\ding{55}} & 1 & - & 1280x720 & 3.2k \\
SkyScapes \cite{azimi2019skyscapes} & 2018 & R & \textcolor{red}{\ding{55}} & Semantic & \textcolor{red}{\ding{55}} & \textcolor{red}{\ding{55}} & \textcolor{red}{\ding{55}} & \textcolor{red}{\ding{55}} & 1 & - & 5616x3744 & 16 \\
UAVid \cite{lyu2020uavid} & 2020 & R & \textcolor{red}{\ding{55}} & Semantic & \textcolor{red}{\ding{55}} & \textcolor{red}{\ding{55}} & \textcolor{red}{\ding{55}} & \textcolor{red}{\ding{55}} & 1 & - & 4096x2160 & 0.4k \\
% WildUav \cite{florea2021wilduav} & 2021 & R & \textcolor{green}{\ding{51}} & Semantic & \textcolor{green}{\ding{51}} & \textcolor{red}{\ding{55}} & \textcolor{red}{\ding{55}} & \textcolor{green}{\ding{51}} & 1 & - & 5280x3956 & 1.5k \\
% VDD \cite{cai2023vdd} & 2023 & R & \textcolor{red}{\ding{55}} & Semantic & \textcolor{red}{\ding{55}} & \textcolor{red}{\ding{55}} & \textcolor{red}{\ding{55}} & \textcolor{red}{\ding{55}} & 1 & - & 4032x3024 & 400 \\
% \midrule
% \multicolumn{13}{l}{\textit{\textbf{Aerial Datasets - Synthetic}}} \\
% MidAir \cite{fonder2019mid} & 2019 & S & \textcolor{green}{\ding{51}} & Semantic & \textcolor{green}{\ding{51}} & \textcolor{red}{\ding{55}} & \textcolor{red}{\ding{55}} & \textcolor{red}{\ding{55}} & 1 & High & 1024x1024 & 119k \\
UrbanScene3D \cite{liqiang2022UrbanScene3D} & 2022 & S/R & \textcolor{green}{\ding{51}} & Instance & \textcolor{green}{\ding{51}} & RGB/Ins & \textcolor{red}{\ding{55}} & \textcolor{green}{\ding{51}} & 1 & - & 8192x5460 & 128k \\
Dronescapes \cite{marcu2023self} & 2023 & R & \textcolor{green}{\ding{51}} & Semantic & \textcolor{green}{\ding{51}} & \textcolor{red}{\ding{55}} & \textcolor{red}{\ding{55}} & \textcolor{red}{\ding{55}} & 1 & - & 3840x2160 &  16.9k \\
Matrix City \cite{li2023matrixcity} & 2023 & S & \textcolor{green}{\ding{51}} & \textcolor{red}{\ding{55}} & \textcolor{green}{\ding{51}} & RGB  & \textcolor{red}{\ding{55}} & \textcolor{green}{\ding{51}} & 1 & Low & 1920x1080 & 410k \\
Syndrone \cite{rizzoli2023syndrone} & 2023 & S & \textcolor{green}{\ding{51}} & Semantic & \textcolor{green}{\ding{51}} & \textcolor{red}{\ding{55}} & \textcolor{red}{\ding{55}} & \textcolor{red}{\ding{55}} & 3 & Medium & 1920x1080 & 72k \\
DDos \cite{kolbeinsson2024ddos} & 2024 & S & \textcolor{green}{\ding{51}} & Semantic & \textcolor{green}{\ding{51}} & \textcolor{red}{\ding{55}} & \textcolor{red}{\ding{55}} & \textcolor{red}{\ding{55}} & 1 & Medium & 1280x720 & 34k \\
\textbf{ClaraVid (ours)} &  & S & \textcolor{green}{\ding{51}} & \textbf{Panoptic} & \textcolor{green}{\ding{51}} & RGB/Sem/Ins & \textcolor{green}{\ding{51}} & \textcolor{green}{\ding{51}} & 3 & \textbf{Low} & 4032x3024 & 16.9k \\
\bottomrule
\end{tabular}
}
\vspace{-0.35cm}
\begin{flushleft}
\footnotesize \textit{*S/R - Synthetic / Real}
\end{flushleft}
\label{tab:datasets}
\vspace{-0.5cm}
\end{table*}
\section{Related work}

\textbf{Aerial datasets} The progress in semantic and geometric scene understanding of urban scenes has been largely driven by multi-modal automotive datasets, real \cite{liao2022kitti, geyer2020a2d2, sun2020scalability} and synthetic \cite{richter2017gta5, wrenninge2018synscapes, gómez2023urbansyn}. These datasets provide strong benchmarks that integrate LiDAR, multi-view imagery, and detailed annotations for tasks enabling autonomous navigation. In contrast, real-world aerial datasets, are often small in scale and fragmented in scope: semantic segmentation \cite{chen2018udd, azimi2019skyscapes, lyu2020uavid, marcu2020semantics, cai2023vdd}, depth estimation \cite{florea2021wilduav, licaret2022ufodepth}, object tracking \cite{li2021uavdt, benchmark_uav_track}, LiDAR segmentation \cite{melekhov2024eclair}, etc. While some works \cite{marcu2023self} exist that integrate both semantic and geometric information, none of them are targeted towards mapping applications. Synthetic aerial datasets, created using graphics engines, offer a scalable alternative with rich semantic content. However, they come with significant limitations. Their resolution is typically limited to 2K \cite{khose2024skyscenes}, whereas real datasets can reach 4K \cite{lyu2020uavid, chen2018udd, cai2023vdd}. Rendering artifacts, such as issues with foliage rendering \cite{fonder2019mid} and inconsistent levels of detail \cite{khose2024skyscenes, rizzoli2023syndrone}, reduce their realism. Additionally, key modalities like semantic segmentation are frequently missing \cite{gómez2023urbansyn, li2023matrixcity, agarwal2023simulating}, making them less effective for semantic mapping applications. \textit{ClaraVid} addresses these limitations directly by providing a dataset specifically designed for mapping tasks. It includes multi-view imagery, panoptic annotations, and dense depth maps, offering a high-fidelity benchmark targeted towards real-world applications. A summary of the presented datasets is provided in \autoref{tab:datasets}.

\textbf{Neural scene representation of urban scenarios from aerial perspective} Recent advancements in neural scene representations have established robust frameworks for modeling complex, real-world environments. NeRFs, introduced by Mildenhall \textit{et al.}\cite{mildenhall2020nerf}, have proven highly effective across a range of tasks--where further extensions shown promising results from novel view synthesis \cite{barron2021mipnerf, mueller2022instantngp} to surface reconstruction \cite{wang2021neus, Wang2023neus2, Oechsle2021UNISURF, li2023neuralangelo} and panoptic reconstruction \cite{kundu2022pnf, siddiqui20233dto2d}. In parallel, Gaussian Splatting \cite{kerbl2023gsplat} has emerged as a particle-based alternative that bypasses the computational overhead of neural networks, thereby enabling significantly accelerated training and rendering for real-time, large-scale scene representations \cite{kerbl2023gsplat,zhou2024hugs,chen2023omnire,kerbl2024hierarchicalgsplat}. However, many of these approaches are relying on prior data for semantic information that can typically be obtained from GT data or as pseudolabels. Finally, while both paradigms have been applied to urban scene reconstruction from aerial perspectives \cite{liu2024citygaussian, turki2022meganerf}, they largely omit semantic information, primarily due to the absence of comprehensive datasets. We introduce our dataset specifically designed to benchmark these diverse methodologies within a unified framework that integrates all relevant modalities from a low aerial viewpoint. 

\textbf{Dataset complexity through the information-theory lens} Differential entropy has emerged as a robust information-theoretic tool for quantifying image complexity, building on Shannon’s definition of entropy\cite{shannon1948mathematical} as a measure of uncertainty in data. Larkin’s formulation of delentropy \cite{larkin2016delentropy} exemplifies this approach, extending traditional entropy to incorporate spatial structure by using a gradient-based joint density (deldensity) that captures pixel co-occurrence and image detail. This enables a more comprehensive complexity measure than conventional metrics: whereas basic Shannon entropy ignores spatial relationships among pixels and GLCM-based texture entropy\cite{haralick1973glcm} captures only limited local patterns, delentropy integrates both local and global features of an image’s intensity distribution. Rahane \textit{et al.}\cite{rahane2020delentropydataset} compared large-scale image datasets using entropy distributions (pixel-level, texture, and delentropy), finding that higher-order measures like delentropy align well with dataset learning difficulty and correlate with known variations in model performance for the semantic segmentation task. Similarly, Cagas \textit{et al.} \cite{cagas2024medicaldelentropy} observed that a dataset’s complexity distribution, quantified via delentropy, correlates with generative model fidelity, informing how increasing image complexity can demand larger training sets for GANs\cite{goodfellow2014generative}. We leverage delentropy to construct a scene-level complexity measure that serves as a proxy for assessing reconstruction difficulty in aerial missions only from 2D images.

\vspace{-0.5cm}
\section{Scene complexity through delentropy}

% overview and task definition
In UAV-based scene reconstruction, we focus on predicting reconstruction complexity directly from 2D aerial imagery, targeting novel view synthesis without access to any reconstructed geometry. This formulation departs from prior work\cite{kim2022infonerf} that computes complexity during or after 3D reconstruction. Instead, our approach estimates complexity a priori from images alone. We restrict the problem to structured mapping scenarios where data acquisition follows grid-based trajectories at discrete altitudes, with bounded spatial extents and minimized peripheral content (e.g., distant backgrounds). These constraints reflect real-world UAV mission protocols and provide a stable foundation for inferring scene difficulty from image characteristics alone. Within this context, we formalize the task of image-based complexity estimation and lay the groundwork for linking it to downstream reconstruction error through a principled  framework.

\subsection{Theoretical background}
We interpret \textbf{image complexity} from an information-theoretic perspective, as the variability of spatial structure encoded within a single view. Classical formulations, such as Shannon's pixel entropy, capture only the intensity distributions, disregarding spatial dependencies. To generalize spatial entropy in a principled manner, we adopt Larkin’s delentropy\cite{larkin2016delentropy}, a reinterpretation of Shannon’s second-order entropy, where entropy is computed over the gradient field. Delentropy can be formalized as follows, given an image \( I \), the gradient components are computed as the partial derivatives:
\begin{equation}
    f_x = \frac{\partial I}{\partial x}, \quad f_y = \frac{\partial I}{\partial y}
\end{equation}
These components represent the rate of change in the \( x \) (vertical) and \( y \) (horizontal) directions, respectively, capturing edges and variations. In practice, these are computed using the Sobel operator\cite{sobel1970sobel}. The joint probability density function, the \emph{deldensity} \( p(f_x, f_y) \) is estimated using a 2D histogram of \( f_x \) and \( f_y \), formally denoted using the Kronecker delta \( \delta \) as follows:
\begin{equation}
    p_{ij} = \frac{1}{4XY} \sum_{n=-N}^{N-1} \sum_{m=-M}^{M-1} \delta_{i,f_x(m,n)} \delta_{j,f_y(m,n)}
\end{equation}
where \( p_{ij} \) represents the joint probability of observing the gradient values (i,j) at any pixel, $N, M$ define the spatial window size over which the gradient distribution is computed and $X, Y$, the image width and height. Delentropy, denoted as \( H_{\text{del}} \), is then defined as the joint entropy of this distribution. If \( p_{ij} \) is the probability of \( f_x \) and \( f_y \) falling into the \( i \)-th and \( j \)-th bins, respectively, the discrete form is:
\begin{equation}
    H_{\text{del}} = -\frac{1}{2} \sum_{i}^{I} \sum_{j}^{J} p_{ij} \log_2 p_{ij}
\end{equation}
such that \(I\) and \(J\) represent the number of bins in the 2D distribution, while the factor \(\tfrac{1}{2}\) is derived from ~\cite{Papoulis1977} and maintained for scale consistency. 

\begin{figure}
    \centering
    \includegraphics[width=\linewidth]{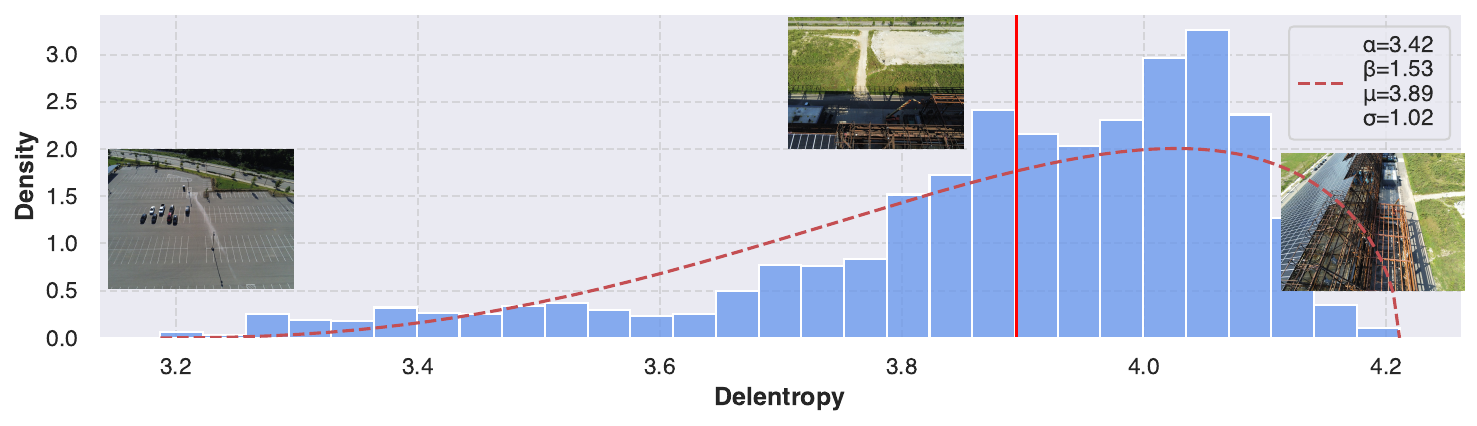}
    \caption{\textbf{Delentropic Scene Profile of Mill19 Building\cite{turki2022meganerf}}. The histogram represents the delentropy distribution, with overlaid sample images illustrating complexity variations. The dashed curve denotes the fitted Beta distribution and the vertical red line marks the mean delentropy.}
    \label{fig:distribution_dsp_example}
    \vspace{-0.45cm}
\end{figure}

% delentropy intuition
This formulation assumes that the spatial distribution of image gradients reflects the underlying structural complexity of the scene. Regions with rich texture, repeated patterns, or fine-grained variations yield broader gradient distributions and thus higher delentropy. Conversely, flat or uniform areas lead to concentrated gradient densities and lower delentropy. This provides a compact measure of spatial complexity directly observable from a single image. A graphical example for different complexity levels is provided in \autoref{fig:distribution_dsp_example}.

\subsection{Delentropic Scene Profile}
% scene profile definition
We define a \textbf{scene complexity profile} as a statistical characterization of the representational difficulty associated with a finite set of images $\{I_k\}_{k=1}^N$ depicting a shared 3D environment. Each image $I_k$ is assigned a scalar complexity measure $C_k \in \mathbb{R}_{\geq 0}$, and the resulting collection $\{C_k\}_{k=1}^N$ induces a distribution that captures the variability of structural content across views. Under structured acquisition protocols (e.g., UAV trajectories with regular grid-based sampling and bounded spatial extent), each image encodes a spatially distinct yet complementary subset of the environment. In such settings, the aggregated complexity distribution offers a principled descriptor of scene-level heterogeneity and serves as a proxy for the intrinsic challenge posed to reconstruction algorithms.

% dsp definition
To instantiate this framework, we adopt \emph{delentropy} $H_{\mathrm{del}}$ as the underlying image complexity measure. For a scene $S = \{I_1, \dots, I_N\}$, we define the corresponding \textbf{Delentropic Scene Profile} (DSP) as the empirical distribution of per-image delentropy values $\{H_{\mathrm{del},k}\}_{k=1}^N$, with $H_{\mathrm{del},k} = H_{\mathrm{del}}(I_k)$. These values are computed over regions deemed relevant for reconstruction (e.g., excluding masked dynamic content). We model this distribution via a truncated Beta distribution parameterized by $(\alpha, \beta, a, b)$:

\begin{align}
\text{DSP}_S &= Beta(H_{\text{del}} \mid \alpha, \beta, a, b) \notag \\
&= \frac{(H_{\text{del}} - a)^{\alpha - 1} (b - H_{\text{del}})^{\beta - 1}}{(b - a)^{\alpha + \beta - 1} B(\alpha, \beta)},
\end{align}
where $B(\alpha, \beta) = \int_0^1 t^{\alpha - 1} (1 - t)^{\beta - 1} \, dt$ is the Beta function, $\alpha, \beta > 0$ are shape parameters, and $a, b$ (with $0 \leq a < b \leq 9.83$) define the support and scale of $H_{\text{del}} \in [a, b]$, consistent with the empirical range of delentropy values of a given scene $S$ given that the Beta distribution is defined over $[0, 1]$. Parameters $(\alpha, \beta, a, b)$ are determined via maximum likelihood estimation, maximizing:
\begin{equation}
L(\alpha, \beta, a, b) = \prod_{k=1}^N f(H_{\text{del},k} \mid \alpha, \beta, a, b).
\end{equation}

\begin{figure*}[ht]
    \centering
    \includegraphics[width=0.98\linewidth]{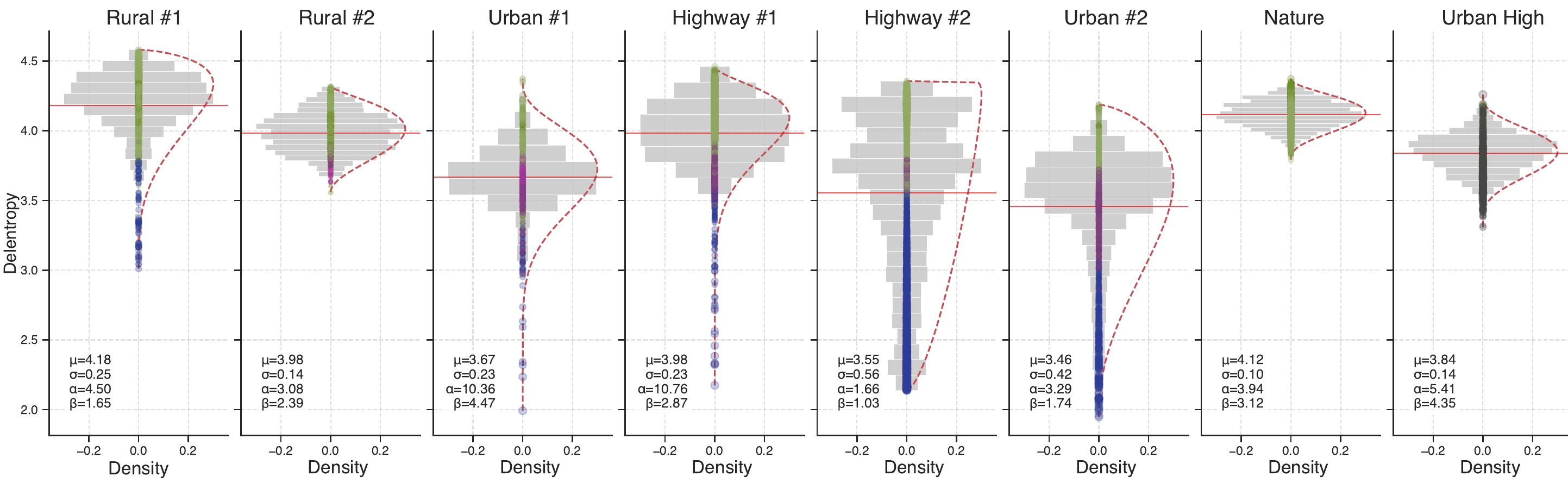}
    \caption{\textbf{ClaraVid Delentropic Scene Profiles} as the distribution of delentropy scores across scenes. Each density plot represents delentropy values per image, with colors indicating the most frequent semantic class. The red horizontal line marks the mean delentropy, while the dashed curve shows the fitted Beta distribution. Water-dominated scenes exhibit long lower tails due to low-texture regions, whereas higher delentropy values, reflecting greater visual complexity, are prevalent in dense vegetation (trees, wheat fields, etc). Statistical parameters ($\mu$, $\sigma$, $\alpha$, $\beta$) quantify scene complexity.}
    \label{fig:delentropy_profiles}
\end{figure*}

\subsection{Delentropic Scene Profile Interpretation}
\textbf{DSP} quantifies scene complexity through four key analytical descriptors: the mean \(\mu\), standard deviation \(\sigma\), and the shape parameters \(\alpha, \beta\) of the Beta-distributed delentropy values, \( H_{\text{del}} \sim \text{Beta}(\alpha, \beta, a, b) \). The mean \(\mu\) encapsulates the overall structural complexity of a scene. Low-complexity environments (\(\mu < 2.5\)) predominantly contain large uniform regions, such as clear skies, bodies of water, or uniformly textured surfaces, whereas high-complexity scenes (\(\mu > 4.75\)) are characterized by high frequency textures such as dense vegetation, roof tilings, etc. The standard deviation \(\sigma\) quantifies the dispersion of delentropy values across the scene, with higher values indicating greater structural heterogeneity. The shape parameters \(\alpha\) and \(\beta\) define the skewness of the delentropy distribution: \(\alpha < \beta\) signifies a predominance of low-detail imagery, while \(\alpha > \beta\) indicates a tendency towards high-complexity views. When \(\alpha \approx \beta < 1\), the distribution is bimodal, capturing scenes with both extreme simplicity and complexity, whereas \(\alpha \approx \beta > 1\) suggests a concentration of delentropy values around moderate complexity levels. Together, these descriptors provide a quantitative framework for characterizing aerial scene complexity, facilitating the selection of optimal reconstruction strategies.

\textbf{Discussion.} In the context of neural scene reconstruction, the DSP provides a lens to interpret different types of predictive uncertainty—epistemic (coverage gaps in training or inherent reconstruction method constraints) and heteroscedastic (irreducible errors). A long tail of low-delentropy images (e.g., large uniform regions such as sky or water) offers minimal geometric cues, intensifying epistemic and model uncertainty. Here, standard strategies like increasing the number of training samples or model size often fail to reduce error; instead, domain-specific solutions—such as incorporating depth priors--have the potential to be more effective\cite{roessle2022dense}. Conversely, a long tail of high-delentropy images, with intricate textures and geometry, (e.g., densely tiled roofs, trees, etc) can drive large variations when the model is under-parameterized, and  additional viewpoints or higher network capacity to capture subtle details. In both extremes cases of delentropy values, heteroscedastic uncertainty limits how much the error can be reduced solely through oversampling or upscaled architectures. \emph{By revealing how complexity is distributed across scenes, the DSP informs targeted resource allocation and methodology choices, that ultimately can improve reconstruction fidelity.}

\textbf{Limitations.} Delentropy inherently relies on image gradients, rendering it sensitive to factors such as blur, resolution variations, or imaging artifacts. In such cases, it becomes unclear whether reconstruction should aim to reproduce the degraded observations or an idealized, high-fidelity version of the scene. Additionally, DSP is formulated under structured acquisition settings, assuming bounded scenes and regular grid-based sampling. When these assumptions break down--for instance, in unbounded environments with significant peripheral content or in scenes requiring spatial contraction--the predictive power of DSP with respect to reconstruction (evaluated via novel view synthesis) quality may deteriorate.

Please consult the Supplementary~\ref{supp:more_on_dsp} for more information regarding design decisions of DSP.
\section{ClaraVid Dataset}

In this section we describe \textit{ClaraVid}, a synthetic dataset designed for semantic and geometric reconstruction tasks. It comprises \textit{16,917 multimodal frames} collected across 5 diverse environments—urban, urban high, rural, highway, and natural landscapes— through 8 distinct UAV missions. ClaraVid spans scenes of varying structural complexity, from uniform environments like Urban or Highway to highly detailed ones such as Rural and Nature. This diversity supports controlled benchmarking across different levels of reconstruction difficulty. We present the delentropic scene profiles in \autoref{fig:delentropy_profiles}. 

\textbf{Data collection.} To ensure comprehensive scene coverage, \textit{ClaraVid} employs a structured grid-based acquisition strategy, capturing images with an 80–90\% overlap to balance redundancy and spatial continuity. Each mission is recorded from 3 distinct viewpoints, simulating a multi-UAV coordination paradigm. Data is acquired at varying altitudes and pitch angles: low (45–55m) 45\textdegree, mid (55–65m) 55\textdegree, and high (65–75m) 90\textdegree, with altitude selected based on terrain complexity. Importantly, frames are captured at 2–3 second intervals, reflecting real-world UAV mapping protocols that prioritize discrete high-quality images over continuous video. \textit{ClaraVid} spans a total area of 1.8 km$^2$, with an average scene coverage of 0.22 km$^2$.

\textbf{Data modalities.} The dataset provides a diverse set of sensor modalities to support neural reconstruction tasks. High-resolution RGB imagery is captured using a resolution of 4032$\times$3024 with a 75$^\circ$ horizontal FoV. To encode geometric priors, \textit{ClaraVid} includes metric depth maps and scene-level sparse point clouds (color, semantic, instance) sampled at three densities (30, 100 and 200 cm), simulating LiDAR scans. For semantic understanding, panoptic segmentation masks provide instance-aware annotations for static structures such as buildings and dynamic entities including vehicles and pedestrians, with a semantic vocabulary of 18 classes. A key feature is the inclusion of dynamic object masks, which isolate moving entities, an essential component for tasks requiring temporal coherence. To the best of our knowledge, this feature is unavailable in prior works. These multimodal annotations position \textit{ClaraVid} as a comprehensive dataset for aerial perception challenges. 

\textbf{Development Environment.} \textit{ClaraVid} is developed using Unreal Engine 4, and leveraging the Carla\cite{dosovitskiy2017carla} ecosystem for sensors and dynamic agent behavior, ensuring diverse traffic patterns. The dataset spans 5 distinct environments. For urban, rural, and highway scenarios, we enhance Carla’s original towns with photorealistic assets, refining textures, lighting, and object placement to elevate scene fidelity beyond default synthetic environments. The natural landscape is created using real-world elevation \cite{IGN_Topographic_Data} and road network priors from OSM\cite{OpenStreetMap}, ensuring topographic realism, while the high-density urban environment is constructed from professionally designed assets that introduce architectural complexity. Across all scenes, procedural generation enriches foliage and ground cover--introducing trees, grass, and natural elements--to counteract the overly uniform, planar compositions found in prior synthetic datasets (\cite{rizzoli2023syndrone, khose2024skyscenes}. Manual refinements ensure spatial coherence and realism, creating environments that balance controlled benchmarking with the visual and structural richness needed for real-world applicability.

Additional information about the dataset can be found in Supplementary~\ref{supp:more_on_claravid}.

\section{Experimental Evaluation}

We conduct a structured evaluation of \emph{ClaraVid} to assess reconstruction and perception tasks across varying scene complexities and viewpoints. First, we benchmark the reconstruction accuracy of multiple neural reconstruction models on complete scene sets, analyzing their performance through \emph{delentropic scene profiling} to quantify the impact of scene complexity on representational capacity. Second, we perform an ablation study on the influence of training viewpoints on multi-modal tasks, including novel view synthesis, semantic segmentation, and depth estimation, to evaluate each model’s adaptability to unseen viewpoints. Finally, we validate the dataset’s utility for downstream tasks by analyzing semantic segmentation results, providing a quantitative assessment of the synthetic-to-real domain gap.

\begin{figure}[t]
    \centering
    \includegraphics[width=\linewidth]{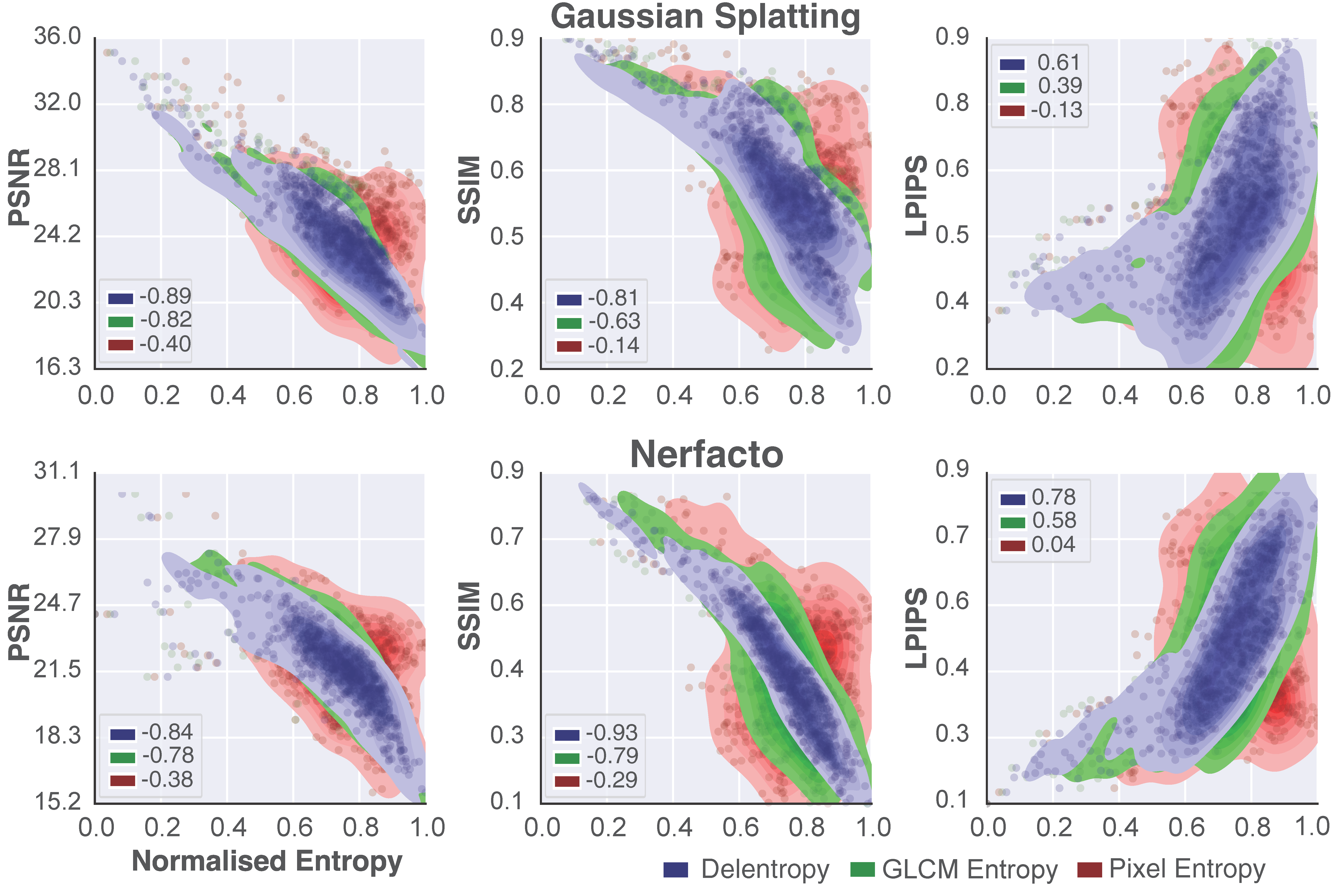}
    \caption{\textbf{Correlation between Complexity Metrics and Reconstruction Error.} Delentropy consistently achieves stronger Pearson correlations and tighter predictive intervals with reconstruction metrics compared to GLCM texture entropy and Shannon pixel entropy. Higher delentropy corresponds directly to increased perceptual errors (LPIPS) and inversely to texture(PSNR) and geometric metrics(SSIM). The results are computed at image level aggregated across test splits.}
    \label{fig:image_delentropy}
\end{figure}

\subsection{Reconstruction Performance}

\begin{table*}[t]
    \centering
    \caption{\textbf{Novel view synthesis evaluation on ClaraVid.} We benchmark several neural reconstruction methods across scenes with different Delentropic Scene Profiles. A higher $\mu$ consistently correlates with lower PSNR/SSIM and higher LPIPS, confirming that increased scene complexity degrades reconstruction quality. While $\alpha$ has limited predictive value, higher $\beta$—indicating skew toward simpler views—partially mitigates performance loss.}
    \renewcommand{\arraystretch}{1.2} 
    \resizebox{\linewidth}{!}{
    \begin{tabular}{l|>{\columncolor{gray!10}}c>{\columncolor{gray!10}}c>{\columncolor{gray!10}}c|ccc|ccc|ccc|ccc|ccc}
        \toprule
        \textbf{Scene} & 
        \multicolumn{3}{c|}{\cellcolor{gray!10} \textbf{DSP}} & 
        \multicolumn{3}{c|}{\textbf{Instant-NGP}\cite{mueller2022instantngp}} & 
        \multicolumn{3}{c|}{\textbf{TensoRF} \cite{chen2022tensorf}} & 
        \multicolumn{3}{c|}{\textbf{Nerfacto}\cite{tancik2023nerfstudio}} & 
        \multicolumn{3}{c|}{\textbf{ZipNeRF}\cite{barron2023zipnerf}} & 
        \multicolumn{3}{c}{\textbf{Gaussian Splatting}\cite{kerbl2023gsplat}} \\
        & \cellcolor{gray!10} \textbf{$\mu$} & \cellcolor{gray!10} \textbf{$\alpha$} & \cellcolor{gray!10} \textbf{$\beta$} 
        & PSNR ↑ & SSIM ↑ & LPIPS ↓ 
        & PSNR ↑ & SSIM ↑ & LPIPS ↓ 
        & PSNR ↑ & SSIM ↑ & LPIPS ↓ 
        & PSNR ↑ & SSIM ↑ & LPIPS ↓ 
        & PSNR ↑ & SSIM ↑ & LPIPS ↓   \\
        \midrule
        Rural \#1    & {\cellcolor{red!45}\textit{4.18}}    & \textit{4.50}  & \textit{1.65} & 18.27 & 0.3171 & 0.8492 & 18.11 & 0.2898 & 0.9204 & 18.36 & 0.2981 & 0.9022 & 18.56 & 0.4462 & 0.8818 & 19.97 & 0.5351 & 0.6019 \\
        Nature       & {\cellcolor{red!25}\textit{4.12}}    & \textit{3.94}  & \textit{3.12} & 21.40 & 0.3993 & 0.7057 & 21.02 & 0.3338 & 0.8213 & 21.60 & 0.3772 & 0.7705 & 24.55 & 0.5184 & 0.5829 & 22.94 & 0.5586 & 0.5515 \\
        Rural \#2    & {\cellcolor{yellow!45}\textit{3.98}} & \textit{3.08}  & \textit{2.39} & 20.75 & 0.3769 & 0.7192 & 20.61 & 0.3525 & 0.7890 & 21.29 & 0.3721 & 0.8145 & 24.54 & 0.5010 & 0.6645 & 22.59 & 0.5402 & 0.5727 \\
        Highway \#1  & {\cellcolor{yellow!25}\textit{3.98}} & \textit{10.76} & \textit{2.87} & 20.32 & 0.3165 & 0.9148 & 20.41 & 0.3240 & 0.9043 & 20.96 & 0.3548 & 0.8862 & 21.73 & 0.4792 & 0.7960 & 22.03 & 0.4769 & 0.6814 \\
        Urban High   & {\cellcolor{lime!25}\textit{3.84}}   & \textit{5.41}  & \textit{4.35} & 21.26 & 0.4878 & 0.5150 & 20.75 & 0.4465 & 0.5564 & 22.39 & 0.4797 & 0.5893 & 24.30 & 0.6920 & 0.4049 & 24.19 & 0.6211 & 0.4324 \\
        Urban \#1    & {\cellcolor{lime!45}\textit{3.67}}   & \textit{10.36} & \textit{4.47} & 22.25 & 0.5507 & 0.4829 & 21.96 & 0.5230 & 0.5271 & 23.24 & 0.5680 & 0.5321 & 26.39 & 0.7451 & 0.3614 & 26.37 & 0.7507 & 0.2981 \\
        Highway \#2  & {\cellcolor{green!25}\textit{3.55}}  & \textit{1.66}  & \textit{1.03} & 22.97 & 0.5174 & 0.6389 & 22.78 & 0.4944 & 0.6887 & 22.66 & 0.5238 & 0.6668 & 24.51 & 0.6964 & 0.4609 & 24.61 & 0.6504 & 0.4858 \\
        Urban \#2    & {\cellcolor{green!45}\textit{3.29}}  & \textit{3.29}  & \textit{1.74} & 23.85 & 0.5734 & 0.4925 & 23.66 & 0.5511 & 0.5438 & 24.24 & 0.5798 & 0.5485 & 25.42 & 0.6945 & 0.3443 & 26.68 & 0.7125 & 0.3658 \\
        \midrule
        \textbf{Average} & \textbf{\textit{3.97}} & \textbf{\textit{5.37}} & \textbf{\textit{2.70}} & 21.38 & 0.4424 & 0.6648 & 21.16 & 0.4144 & 0.7189 & 21.84 & 0.4442 & 0.7138 & \textbf{23.75} & 0.5966 & 0.5622 & 23.67 & \textbf{0.6057} & \textbf{0.4987} \\
        \bottomrule
    \end{tabular}%
    }
    \label{tab:novel_view}
    \vspace{-0.35cm}
\end{table*}

We evaluate the reconstruction performance of various neural reconstruction models--\textit{Nerfacto} \cite{tancik2023nerfstudio}, \textit{TensoRF} \cite{chen2022tensorf}, \textit{Instant-NGP} \cite{mueller2022instantngp}, \textit{ZipNeRF} \cite{barron2023zipnerf}, and \textit{Gaussian Splatting} \cite{kerbl2023gsplat}--on the ClaraVid dataset, covering diverse scenarios characterized by distinct \textbf{DSP}. Performance is evaluated using PSNR\cite{gonzalez2009psnr}, SSIM\cite{wang2004ssim}, and LPIPS\cite{zhang2018lpips}, with results summarized in \autoref{tab:novel_view}. Full experimental setup can be found in Supplementary ~\ref{supp:recon_perf_annex}. A clear trend emerges where increased scene complexity, indicated by higher DSP values, leads to a consistent decline in reconstruction fidelity. This is reflected in the strong negative correlation between PSNR/SSIM and DSP mean values, alongside a positive Pearson correlation with LPIPS. A higher $\beta$ appears to mitigate reconstruction degradation, suggesting that scenes with simpler images are less susceptible to performance drops. Among all methods, \textit{Gaussian Splatting} and \textit{ZipNeRF} deliver the most robust performance across settings.

\subsection{DSP as complexity proxy}

To evaluate delentropy as a proxy of reconstruction difficulty, we analyze its correlation with reconstruction error across two  models--\textit{Gaussian Splatting} and \textit{Nerfacto}--on the ClaraVid dataset. Following the setup from the previous subsection, \textbf{DSP} is computed on grayscale frames prior to training, excluding masked dynamic regions. As shown in \autoref{fig:image_delentropy}, delentropy achieves consistently higher correlation with reconstruction metrics than classical complexity measures such as \textit{Shannon entropy}\cite{shannon1948mathematical} and \textit{GLCM texture entropy}\cite{haralick1973glcm}. It correlates negatively with PSNR and SSIM, and positively with LPIPS, indicating higher reconstruction errors in structurally complex regions. This trend is consistent across both models, suggesting that delentropy captures structural cues relevant to reconstruction regardless of model class. Moreover, \textit{delentropy} yields tighter predictive intervals, validating its reliability as a proxy for reconstruction complexity and motivating its use in the Delentropic Scene Profile.

\subsection{Performance Across Varying Viewpoints}

\begin{figure*}[ht]
    \centering
    \includegraphics[width=\linewidth]{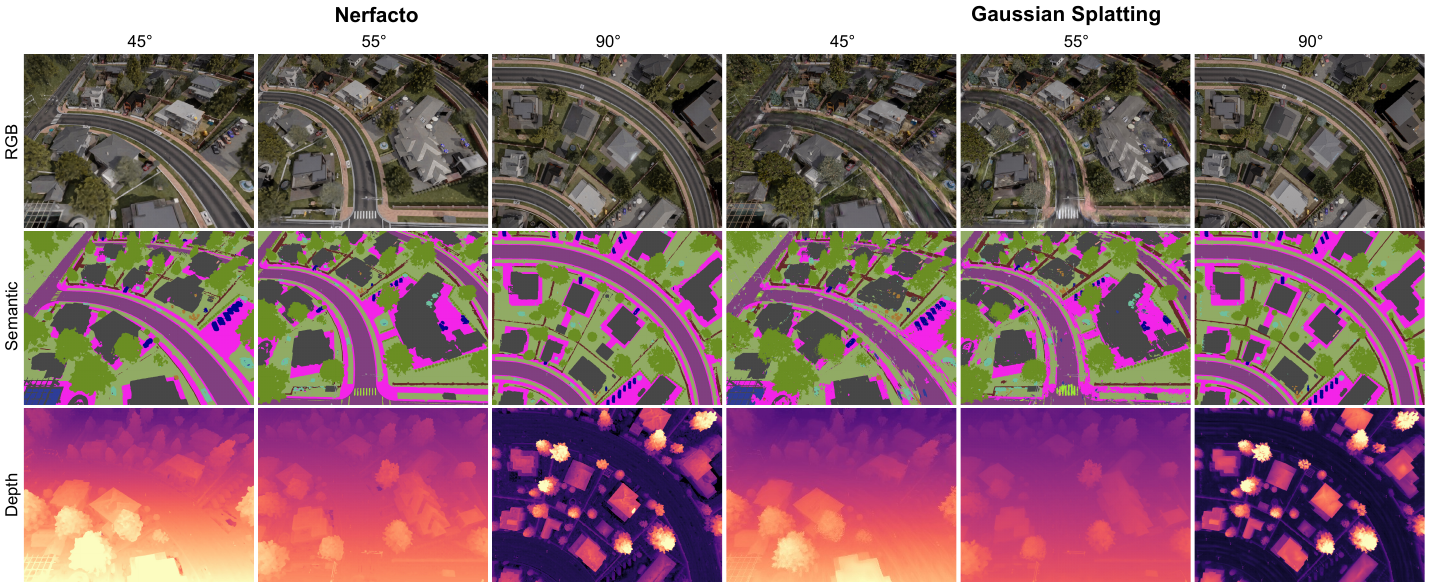}
    \caption{\textbf{Viewpoint Generalization in Neural Reconstruction.} 
    Comparison of Nerfacto and \textbf{3DGS} across RGB, semantics, and depth. Trained on 90\textdegree$|$\textbf{H}, Nerfacto generalizes better to unseen views (45\textdegree$|$\textbf{L}, 55\textdegree$|$\textbf{M}), while 3DGS excels in depth but degrades under shifts. Semantic accuracy follows reconstruction trends, with 3DGS performing well at training views but losing consistency under extrapolation.}
    \label{fig:novel_view}
    \vspace{-0.25cm}
\end{figure*}

\begin{table}[t]
\centering
\caption{\textbf{Performance Across Varying Viewpoints} . Evaluation on appearance, segmentation, and depth synthesis across various viewpoints: \( 45^\circ |\text{\textbf{L}ow} \), \( 55^\circ |\text{\textbf{M}id} \), and \( 90^\circ | \text{\textbf{H}igh} \). (pitch$|$altitude)}
\resizebox{0.84\linewidth}{!}{%
\begin{tabular}{l|ccc|ccc}
\toprule
 & \multicolumn{3}{c}{\textbf{Nerfacto}\cite{tancik2023nerfstudio}} & \multicolumn{3}{c}{\textbf{Gaussian Splatting}\cite{kerbl2023gsplat}} \\
\cmidrule(lr){2-4} \cmidrule(lr){5-7}
\textbf{Train \textbackslash Pred} & $45^\circ$ L & $55^\circ$ M & $90^\circ$ H & $45^\circ$ L & $55^\circ$ M & $90^\circ$ H \\
\midrule
\rowcolor{gray!10} \multicolumn{7}{l}{\textbf{Appearance Synthesis}} \\
\midrule
\multicolumn{7}{l}{\textbf{PSNR} $\uparrow$} \\
$45^\circ$ L  & \textbf{21.36}  & 20.46   & 19.08   & \textbf{22.98}  & 21.56  & 20.50 \\
$55^\circ$ M  & 20.25  & \textbf{21.43}   & 20.48   & 20.75  & \textbf{22.75}  & 21.06 \\
$90^\circ$ H  & 18.23  & 19.42   & \textbf{22.45}   & 17.88  & 19.26   & \textbf{23.00} \\
\midrule
\multicolumn{7}{l}{\textbf{SSIM} $\uparrow$} \\
$45^\circ$ L  & \textbf{0.42}   & 0.40    & 0.36    & \textbf{0.58}   & 0.50   & 0.42  \\
$55^\circ$ M  & 0.39   & \textbf{0.42}    & 0.40    & 0.45  & \textbf{0.58}   & 0.41  \\
$90^\circ$ H  & 0.34   & 0.40    & \textbf{0.47}    & 0.36   & \textbf{0.65}   & 0.59  \\
\midrule
\multicolumn{7}{l}{\textbf{LPIPS} $\downarrow$} \\
$45^\circ$ L  & 0.75   & \textbf{0.74}    & 0.76   & \textbf{0.55}   & 0.61   & 0.78  \\
$55^\circ$ M  & 0.77   & \textbf{0.73}   & 0.74    & 0.59   & \textbf{0.53}   & 0.62  \\
$90^\circ$ H  & 0.79   & 0.71   & \textbf{0.68}    & 0.81   & 0.73   & \textbf{0.50}  \\
\midrule
\rowcolor{gray!10} \multicolumn{7}{l}{\textbf{Semantic Segmentation}} \\
\midrule
\multicolumn{7}{l}{\textbf{mIoU} $\uparrow$} \\
$45^\circ$ L &\textbf{44.82} & 44.14 & 40.78 & \textbf{56.53} & 55.11 & 53.50 \\
$55^\circ$ M & 43.12 & \textbf{46.57} & 45.74 & 53.34 & \textbf{59.59} & 56.50 \\
$90^\circ$ H & 36.95 & 41.64 & \textbf{51.08} & 36.68 & 40.82 & \textbf{58.54} \\
\midrule
\rowcolor{gray!10} \multicolumn{7}{l}{\textbf{Depth}} \\
\midrule
\multicolumn{7}{l}{\textbf{Abs Rel} $\downarrow$} \\
$45^\circ$ L & 0.059 & \textbf{0.042} & 0.066 & 0.025 & 0.017 & \textbf{0.010} \\
$55^\circ$ M & 0.042 & 0.043 & \textbf{0.027} & 0.023 & 0.016 & \textbf{0.009} \\
$90^\circ$ H & 0.054 & 0.039 & \textbf{0.022} & 0.028 & 0.019 & \textbf{0.010} \\
\midrule
\multicolumn{7}{l}{\textbf{RMSE} $\downarrow$} \\
$45^\circ$ L & 12.34  & \textbf{10.64}  & 14.22  & 4.67  & 3.37  & \textbf{2.36} \\
$55^\circ$ M & 11.47  & 9.09   & \textbf{6.55}   & 4.61  & 3.07  & \textbf{1.93} \\
$90^\circ$ H & 14.20  & 8.95   & \textbf{4.68}   & 5.41  & 3.53  & \textbf{2.08} \\
\midrule
\multicolumn{7}{l}{\textbf{$\boldsymbol{\delta < 1.05}$} $\uparrow$} \\
$45^\circ$ L & 0.8193 & \textbf{0.8467} & 0.8317 & 0.881 & 0.92  & \textbf{0.95} \\
$55^\circ$ M & 0.8315 & 0.8584 & \textbf{0.9024} & 0.895 & 0.93  & \textbf{0.96} \\
$90^\circ$ H & 0.7886 & 0.8354 & \textbf{0.8980} & 0.871 & 0.91  & \textbf{0.95} \\
\bottomrule
\end{tabular}%
}
\vspace{-0.4cm}
\label{tab:ablation_novel_view}
\end{table}

We evaluate two neural reconstruction paradigms—NeRF-based \textit{Nerfacto} and particle-based \textit{Gaussian Splatting}--in aerial viewpoint generalization. Motivated by real-world UAV mapping scenarios requiring flexible viewpoint usage, each method is trained at a fixed altitude angle ($45^\circ$, $55^\circ$, or $90^\circ$) and tested on unseen views. Quantitative results --see \autoref{tab:ablation_novel_view}-- highlight distinct trade-offs: Nerfacto demonstrates robust generalization across novel views at the expense of longer training times, while Gaussian Splatting offers superior reconstruction fidelity and rapid convergence but deteriorates markedly with viewpoint shifts. Semantic segmentation (mIoU) mirrors reconstruction trends, with Gaussian Splatting excelling at training angles yet faltering under extrapolation, indicating sensitivity to viewpoint consistency. Conversely, Gaussian Splatting consistently outperforms Nerfacto in depth estimation (RMSE, AbsRel, $\delta^{1.05}$), reflecting its superior geometric accuracy. We used a stricter threshold of $\delta < 1.05$, as even minor inaccuracies can propagate into large errors at great distances ($>100m$). Qualitative results are presented in ~\autoref{fig:novel_view}.

\begin{table}[t]
\centering
\caption{\textbf{Performance on semantic segmentation}. Entropy refers to the normalized Shannon for measuring prediction confidence.}
\label{tab:synth_to_real}
\resizebox{0.85\linewidth}{!}{%
\begin{tabular}{llrr}
\toprule
\textbf{Training} & \textbf{Testing} & \textbf{Entropy} $\down$ & \textbf{mIoU} $\up$ \\
\midrule
\rowcolor{gray!10} \multicolumn{4}{l}{\textit{Oracle/Baseline}} \\
\midrule
UAVid\cite{lyu2020uavid} & UAVid & 0.144 & 71.36 \\
UDD\cite{chen2018udd} & UDD & 0.047 & 79.84 \\
SkyScapes\cite{azimi2019skyscapes} & SkyScapes & 0.176 & 65.49 \\
ClaraVid & ClaraVid & 0.095 & 82.51 \\
Syndrone & Syndrone & 0.051 & 86.36 \\
Dronescapes & Dronescapes & 0.153 & 52.74 \\

\midrule
\rowcolor{gray!10} \multicolumn{4}{l}{\textit{Synthetic to Real}} \\
\midrule
ClaraVid & UAVid & 0.213 & 57.50 \\
Syndrone & UAVid & 0.235 & 52.49 \\
ClaraVid & UDD & 0.196 & 58.22 \\
Syndrone & UDD & 0.213 & 54.43 \\
ClaraVid & SkyScapes & 0.290 & 47.52 \\
Syndrone & SkyScapes & 0.310 & 42.75  \\
ClaraVid & Dronescapes & 0.357 & 37.30  \\
Syndrone & Dronescapes & 0.345 & 34.11  \\

\bottomrule
\end{tabular}
}
\label{tab:synthtoreal}
\vspace{-0.4cm}
\end{table}
\subsection{Semantic Segmentation as Downstream Task}
We assess ClaraVid’s effectiveness in downstream tasks by quantifying the synt-to-real domain gap in semantic segmentation. A DPT-based model is trained on real \cite{lyu2020uavid, chen2018udd, azimi2019skyscapes, marcu2023self} and synthetic datasets \cite{rizzoli2023syndrone}, with results shown in \autoref{tab:synth_to_real}. Despite containing 4.5$\times$ fewer samples than Syndrone \cite{rizzoli2023syndrone}, ClaraVid achieves a smaller domain gap, demonstrating superior scene complexity and image fidelity. Models trained on ClaraVid consistently outperform those trained on Syndrone across all real datasets, achieving higher mIoU and lower prediction entropy. However, a residual semantic gap remains, with ClaraVid exhibiting an average gap of 17.82\%, compared to 22.34\% for Syndrone.

\section{Conclusions}

In this work, we introduced ClaraVid, a synthetic aerial dataset designed to bridge the gap in holistic scene reconstruction tailored for UAV-based scene understanding. We further proposed the Delentropic Scene Profile, a novel complexity metric rooted in differential entropy, which quantifies scene difficulty and exhibits strong correlations with reconstruction performance. Our benchmarking framework systematically evaluates neural reconstruction techniques, revealing how DSP can be a reliable predictor of reconstruction accuracy. Through extensive experimentation, we demonstrate that ClaraVid not only surpasses existing synthetic datasets in realism and annotation quality but also generalizes better to real-world UAV imagery, challenging the conventional notion that dataset size alone dictates generalization. By publicly releasing ClaraVid, we aim to advance UAV-based perception and improve benchmarking for semantic and geometric reconstruction.

\section{Acknowledgment}
This work was supported by the project "Romanian Hub for Artificial Intelligence-HRIA", Smart Growth, Digitization and Financial Instruments Program, MySMIS no. 334906.

{
    \small
    \bibliographystyle{unsrt}
    \bibliography{main}
}

\clearpage
\setcounter{page}{1}
\maketitlesupplementary

\section{More on Delentropic Scene Profile}
\label{supp:more_on_dsp}
This section expands on the design and interpretation of the Delentropic Scene Profile. We justify the use of the Beta distribution and clarify the conditions required for reliable cross-scene comparison. These clarifications aim to guide appropriate usage of DSP as a diagnostic tool in structured 3D scene understanding.

\subsection{Motivation for Beta Distribution} 
\label{supp:motivation_beta_distrib}
The choice of distributional model for $\text{DSP}_S$ was guided by the analysis of  complexity profiles of diverse datasets—ClaraVid, Skydrone\cite{rizzoli2023syndrone}, Skyscapes\cite{azimi2019skyscapes}, UAVid\cite{lyu2020uavid}, 3D Matrix City \cite{li2023matrixcity}, Mill19 \cite{turki2022meganerf}, and UrbanScene3D \cite{liqiang2022UrbanScene3D}—observing varied distributional forms contingent on the reconstruction context. A Gaussian distribution, while apt for uniformly complex scenes, fails to capture skewed or long-tailed distributions prevalent in heterogeneous environments (for example where we have a body of water or sky regions). Alternatives such as the \textit{Pareto} or \textit{Gamma} distributions accommodate skewness but lack flexibility for U-shaped profiles, which emerge in scenarios with pronounced bimodal complexity (e.g., a field with occasional high-detail trees). The Beta distribution, however, adeptly models this spectrum—uniform, skewed, and U-shaped—owing to its shape parameters $\alpha$ and $\beta$, requiring only four parameters (including bounds $a$ and $b$) for a parsimonious yet expressive fit. This simplicity affords intuitive interpretation: $\alpha$ and $\beta$ directly govern the density’s form, while $a$ and $b$ anchor it to the empirical delentropy range. Multimodal distributions were eschewed, as their complexity—necessitating mixture models with additional parameters—hampers interpretability, risks overfitting, particularly given the finite sample sizes of $\{ H_{\text{del},k} \}$. The Beta distribution emerges as the best choice, balancing flexibility, robustness, and theoretical clarity.

\subsection{Comparability Across Scenes}
\label{supp:dsp_across_scenes}
The Delentropic Scene Profile provides a statistical summary of structural complexity and is designed to support comparisons of scenes in terms of their representational difficulty. While delentropy provides an absolute measure of complexity, its reliability for cross-scene comparison hinges on controlled acquisition conditions. \emph{Image resolution} plays a critical role: small variations (e.g., $\approx$10\%) tend to preserve the stability of the DSP, whereas more aggressive change of scales (e.g., $\geq$25\%) can significantly alter high-frequency content, shifting the delentropy distribution (e.g., roof tiles may become indistinct when downsampled). The \emph{spatial extent} of the covered scene is similarly influential. We have observed that DSP remains stable even under 4× scaling in area, but larger deviations or differences in semantic composition introduce shifts in the aggregated profile due to increased structural heterogeneity. Furthermore, the \emph{image collection policy}--particularly the trajectory and sampling layout--can induce biases in the captured content, making structurally similar scenes appear dissimilar under mismatched acquisition patterns. These observations emphasize that to fully leverage DSP as a comparative tool, one should ensure alignment in resolution, spatial coverage, and collection policy—allowing differences in DSP to reflect genuine variations in scene complexity rather than artefacts of sampling.

\subsection{Implementation Details}
The delentropy of an image is obtained by first applying Gaussian blur to the input image using a spatial kernel of $3 \times 3$ pixels with a standard deviation of $1.0$, reducing high-frequency noise and mitigating sensitivity to minor textural variations. Subsequently, spatial gradients along horizontal and vertical directions are computed via Sobel filters with kernel size $3 \times 3$. The resulting gradient field is quantized into a two-dimensional histogram, the \textit{deldensity}, employing $256$ bins per axis. This choice is driven by practical considerations to avoid histogram saturation, an artifact typically encountered when considering both positive and negative gradient values. Finally, the normalized deldensity is used as the joint probability distribution, from which delentropy is calculated.

\section{Supplementary Evaluation for DSP}
\label{supp:supp_eval_for_dsp}
In this section, we evaluate the \emph{DSP} on real-world datasets to assess its robustness under varied capture policies and reconstruction conditions.

\subsection{UAVid}
\label{supp:uavid_eval}
To evaluate the transferability and generalization of the DSP to real-world UAV imagery, we assess its correlation with reconstruction accuracy on the UAVid dataset. Unlike ClaraVid, UAVid features continuous image sequences captured under linear, L-shaped, and U-shaped trajectories, with no explicit grid-based coverage. We select six scenes (seq\_13, seq\_15, seq\_29, seq\_31, seq\_36, seq\_38), prioritizing static intervals. Full experimental setup provided in Section ~\ref{supp:uavid_eval_setup}.

\begin{figure}[t]
    \centering
    \includegraphics[width=\linewidth]{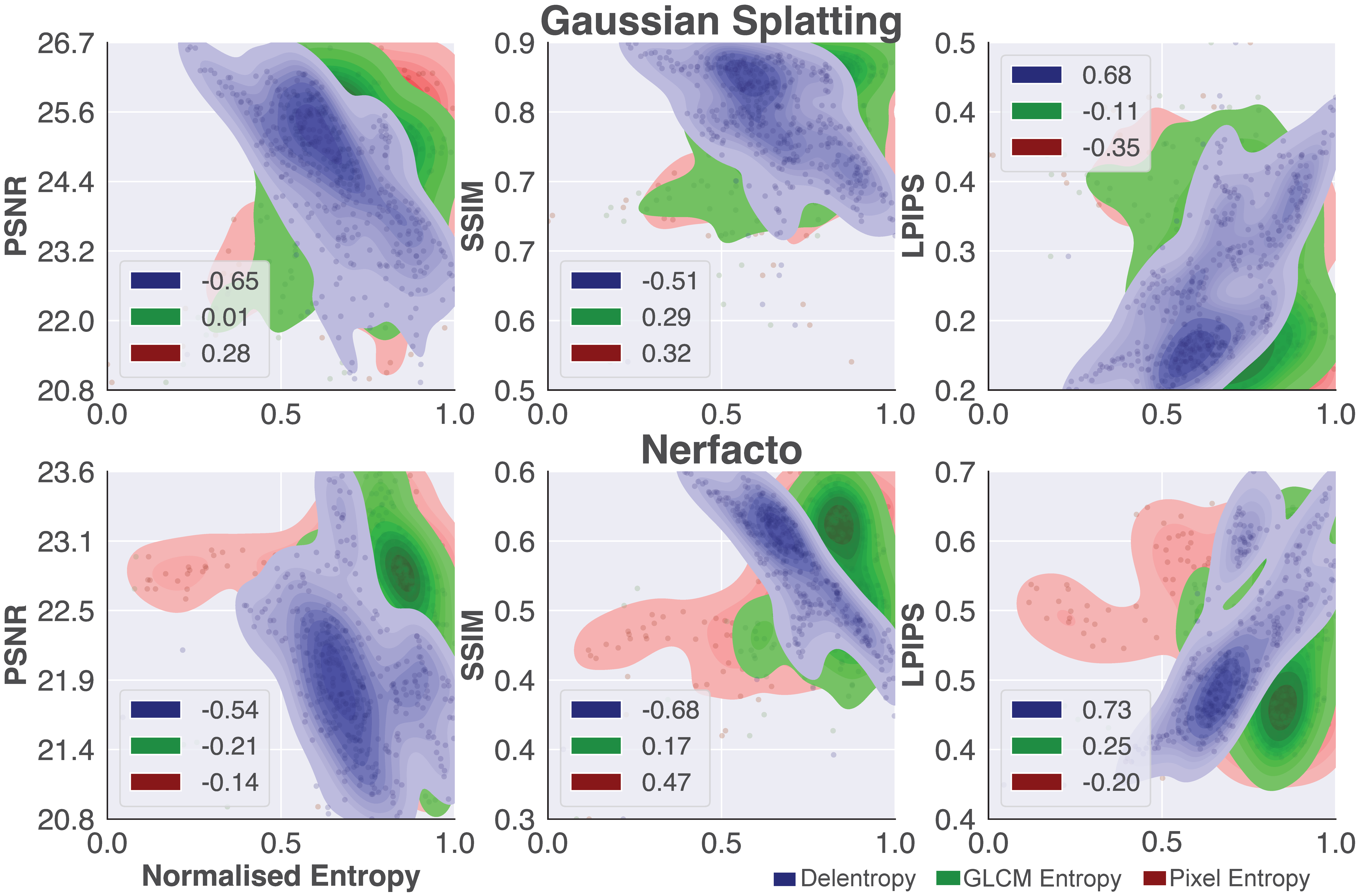}
    \caption{\textbf{Correlation between complexity metrics and reconstruction quality on UAVid.}
    Delentropy (blue) exhibits the strongest and most consistent correlation with PSNR, SSIM, and LPIPS across both evaluated models, outperforming GLCM texture entropy (green) and Shannon pixel entropy (red). Metrics are computed per image and aggregated over the test split.}
    
    \label{fig:supp_uavid_eval_corr}
\end{figure}

\begin{figure*}[t]
    \centering
    \includegraphics[width=\linewidth]{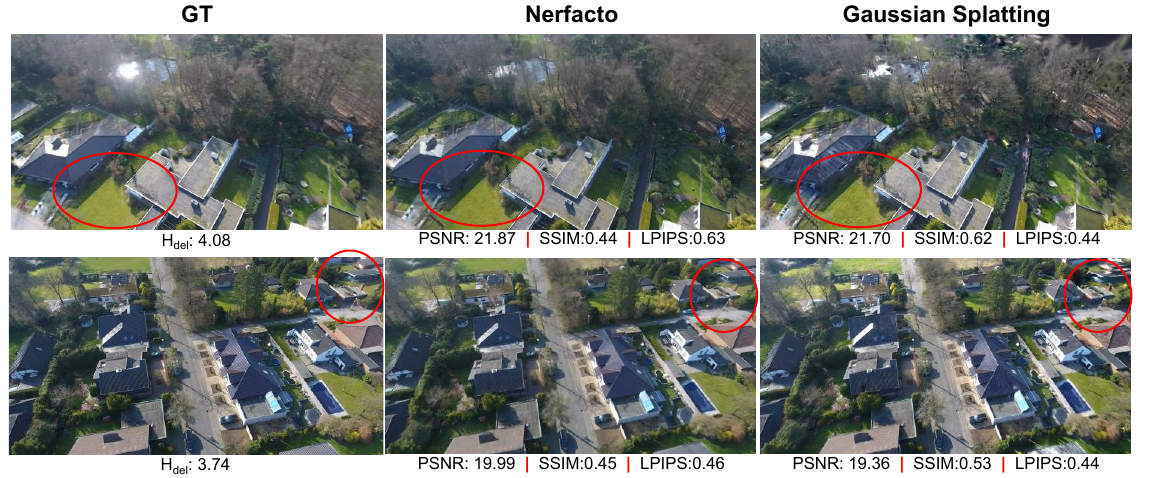}
    \caption{\textbf{Failure cases.} \textbf{Top:} Despite low delentropy and detailed Gaussian Splatting reconstruction, large errors arise from poor camera registration in the blurred input. \textbf{Bottom:} Incomplete scene coverage degrades performance; additionally, the ground truth image exhibits mild rolling shutter distortion.}
    \label{fig:dsp_failure_uavid}
    \vspace{-0.3cm}
\end{figure*}

Despite a reduction in correlation between DSP and reconstruction metrics, the results--summarized in \autoref{tab:uavid_real_eval}--remain consistent with prior trends observed in synthetic settings as presented in \autoref{fig:supp_uavid_eval_corr}. Notable degradations are attributed to real-world artifacts such as rolling shutter, motion blur, and the absence of structured spatial coverage, which collectively weaken the delentropy signal. We identify two recurring failure modes: (i) blurred input frames that yield reconstructions of unexpectedly high fidelity, and (ii) insufficient coverage in low-altitude flyovers (e.g., single-pass views) that lead to poor reconstructions despite an average measured delentropy. Visual examples provided in \autoref{fig:dsp_failure_uavid}.

\begin{table}[t]
    \centering
    \caption{\textbf{DSP evaluation on UAVid.} Higher DSP $\mu$ correlates with lower PSNR, SSIM and higher LPIPS.}
    \renewcommand{\arraystretch}{1.2}
    \resizebox{\linewidth}{!}{%
    \begin{tabular}{l|>{\columncolor{gray!10}}c>{\columncolor{gray!10}}c>{\columncolor{gray!10}}c|ccc|ccc}
        \toprule
        \textbf{Scene} &
        \multicolumn{3}{c|}{\cellcolor{gray!10}\textbf{DSP}} &
        \multicolumn{3}{c|}{\textbf{Nerfacto}\cite{tancik2023nerfstudio}} &
        \multicolumn{3}{c}{\textbf{Gaussian Splatting} \cite{kerbl2023gsplat}}\\
        & \cellcolor{gray!10}\textbf{$\mu$} & \cellcolor{gray!10}\textbf{$\alpha$} & \cellcolor{gray!10}\textbf{$\beta$}
        & PSNR ↑ & SSIM ↑ & LPIPS ↓
        & PSNR ↑ & SSIM ↑ & LPIPS ↓\\
        \midrule
        seq 38 & {\cellcolor{green!25}\textit{3.984}} & \textit{6.147} & \textit{2.216} & 22.83 & 0.637 & 0.536 & 25.18 & 0.857 & 0.234\\
        seq 36 & {\cellcolor{lime!25}\textit{3.990}} & \textit{1.264} & \textit{0.812} & 22.80 & 0.543 & 0.493 & 25.10 & 0.818 & 0.215\\
        seq 29 & {\cellcolor{yellow!25}\textit{4.017}} & \textit{3.187} & \textit{1.702} & 21.93 & 0.525 & 0.512 & 24.07 & 0.770 & 0.238\\
        seq 13 & {\cellcolor{yellow!25}\textit{4.018}} & \textit{6.911} & \textit{2.159} & 21.41 & 0.497 & 0.578 & 23.70 & 0.730 & 0.326\\
        seq 31 & {\cellcolor{orange!25}\textit{4.033}} & \textit{1.470} & \textit{1.659} & 21.75 & 0.463 & 0.475 & 23.93 & 0.718 & 0.191\\
        seq 15 & {\cellcolor{red!25}\textit{4.048}} & \textit{9.008} & \textit{2.300} & 21.20 & 0.487 & 0.547 & 22.98 & 0.709 & 0.350\\
        \bottomrule
    \end{tabular}}
    \label{tab:uavid_real_eval}
    \vspace{-0.5cm}
\end{table}

\subsection{DSP Analysis on Large Reconstructions}

\label{supp:dsp_prior_real_world}
\begin{figure}[h]
    \centering
    \includegraphics[width=\linewidth]{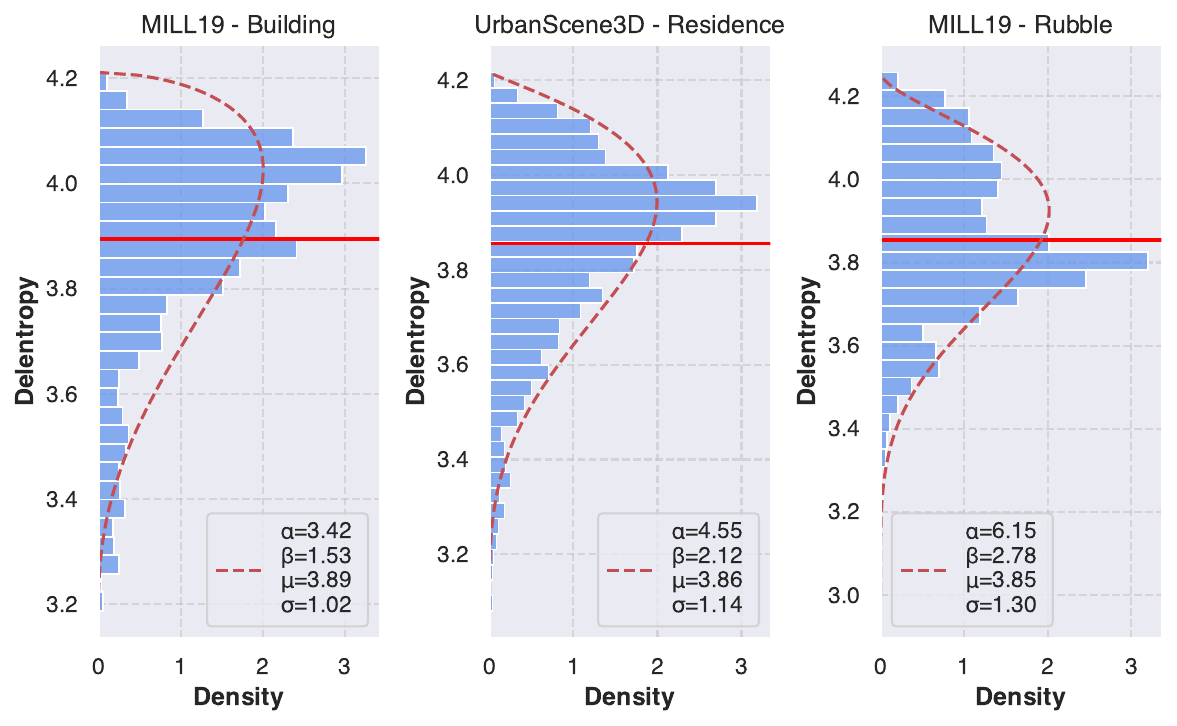}
    \caption{\textbf{Delentropy profiles for real-world mapping datasets}. The DSP indicates medium complexity, with variations reflecting structural differences across scenes. The profile is computed across all scene images.}
    \label{fig:dsp-analysis-real}
    \vspace{-0.3cm}
\end{figure}

To further assess the generalization of \emph{DSP} in real-world conditions, we analyze datasets used in recent large-scale urban reconstruction works, following the experimental setup of CityGaussian~\cite{liu2024citygaussian}. Specifically, we compute DSP values on the training and evaluation splits of UrbanScene3D~\cite{liqiang2022UrbanScene3D} and Mill19~\cite{turki2022meganerf}, which capture diverse urban geometries under unconstrained conditions. These datasets include varying levels of structural complexity, from sparse residential layouts to dense industrial sites. We align our computed DSP statistics--presented in \autoref{fig:dsp-analysis-real}--with published reconstruction scores reported in CityGaussian, enabling a post-hoc correlation analysis without re-running the experiments.

\begin{figure}[h]
    \centering
    \includegraphics[width=\linewidth]{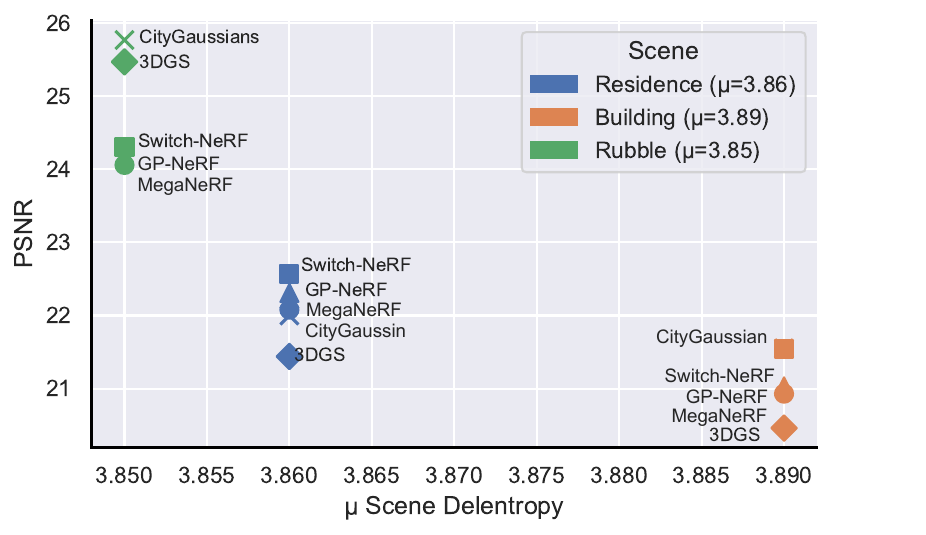}
    \caption{\textbf{Reconstruction quality vs. scene complexity}. The relationship between scene complexity, measured by the $\mu$ mean delentropy, and reconstruction fidelity (PSNR) across different neural reconstruction methods. A higher delentropy value correspond to lower PSNR scores, indicating increased reconstruction difficulty. }
    \label{fig:del_real_methods}
\end{figure}

As illustrated in \autoref{fig:del_real_methods}, we observe a general trend whereby scenes with higher DSP values correspond to lower reconstruction quality, particularly in PSNR. For example, the Mill19 industrial scene exhibits elevated delentropy ($\mu$ $> 3.8$) and lower PSNR across multiple reconstruction methods, comparable to Urban High scenes in ClaraVid. This negative correlation persists across architectures--Mega-NeRF~\cite{turki2022meganerf}, CityGaussian~\cite{liu2024citygaussian}, GN-NeRF~\cite{li2024gpnerf}, Switch-NeRF~\cite{mi2023switchnerf}, and Gaussian Splatting~\cite{kerbl2023gsplat}--suggesting that DSP offers a dataset-agnostic signal of structural difficulty. SSIM and LPIPS exhibited more variability, and their correlation with DSP was less pronounced than that of PSNR. The consistency of trends across methods and environments reinforces DSP’s utility as a lightweight proxy for estimating reconstruction difficulty in real-world benchmarks.

\section{Experimental Setup}
\label{supp:experimental_setup}

\subsection{Reconstruction Performance}
\label{supp:recon_perf_annex}
Throughout our experiments, we utilize the \textit{Nerfstudio} framework\cite{tancik2023nerfstudio} to systematically benchmark various neural reconstruction methods. For \textbf{Nerfacto}\cite{tancik2023nerfstudio}, we adopt the \textit{big} configuration. \textbf{TensoRF}\cite{chen2022tensorf} is configured with an increased grid resolution of $500^3$. \textbf{InstantNGP}\cite{mueller2022instantngp} employs 16 grid levels, a maximum resolution of 8192, and a hash map size of $2^{21}$. \textbf{Zip-NeRF}\cite{barron2023zipnerf} is re-implemented and used with  default configuration. \textbf{Gaussian Splatting}\cite{kerbl2023gsplat} is initialized from the dataset's scene-level point cloud with a spatial resolution of \textit{100 cm}, restricting point splitting beyond 3 million points to balance memory efficiency and reconstruction accuracy. Training durations vary across models, ranging from 60 to 720 minutes on an NVIDIA A6000 GPU, depending on architectural complexity. Across all experiments, we leverage FP16 training to optimize memory usage. The reported reconstruction and segmentation results correspond to a half-resolution output relative to the original image dimensions, maintaining a trade-off between computational feasibility and fidelity in large-scale aerial scene reconstruction.

\subsection{Performance Across Varying Viewpoints}
\label{supp:perf_across_viewpoints}
Both Nerfacto and Gaussian Splatting use the configuration described previously in  Supplementary~\ref{supp:recon_perf_annex}. For semantic learning, Nerfacto incorporates a jointly trained segmentation field, while Gaussian Splatting learns a separate set of 32-width semantic parameters—trained jointly but independent of geometry or positional encodings. Additionally, a 2-layer CNN classifier refines segmentation on the Gaussian splats. We compute per scene class weighting to balance semantic segmentation classes. The depth for both methods is obtained from rendering with no additional heads or learned parameters.

\subsection{Semantic Segmentation}
\label{supp:semantic_seg}
We train a DPT\cite{ranftl2021dpt} model with a DinoV2-L\cite{tian2024dinov2} backbone on both synthetic and real-world UAV datasets. The backbone is kept frozen while only the decoder is optimized. Our training protocol is standardized across datasets, with the number of steps adjusted proportionally to the dataset size—ranging from 5,000 to 30,000—to prevent overfitting. We use each dataset’s native resolution and perform random crops of 630×630 pixels. Additionally, we uniformly apply data augmentations such as color adjustments, lighting variations, rotations, and distortions. Finally, the semantic palette is adapted to the real dataset following the \cite{rizzoli2023syndrone} protocol, where both training and evaluation are done using reduced pallette. 

\subsection{UAVid Evaluation}
\label{supp:uavid_eval_setup}
For the UAVid evaluation, we focus on mostly static scenes, as dynamic object masks are not available. We select six representative sequences (13, 15, 29, 31, 36, 38), each covering an area of approximately $\approx$ 0.1 km$^2$. All sequences consist of 901 frames, training samples are selected at every fourth index, while testing ones are chosen at every second offset (i.e., $i \equiv 2 \pmod{4}$). Stationary frames with minimal motion are excluded, as they often are poorly registered. Additionally, we consider the first and last 100 frames in train set, to mitigate any potential collection policy errors and minimize the spatial content outside the region of interest. All images are downsampled by a factor of 2, and reconstructions are performed using COLMAP\cite{schoenberger2016sfm, schoenberger2016mvs}. All other settings for the evaluated models follow the protocol described in Section~\ref{supp:recon_perf_annex}.

\section{Claravid}
\label{supp:more_on_claravid}
\begin{figure*}[ht]
    \centering
    \includegraphics[width=\linewidth]{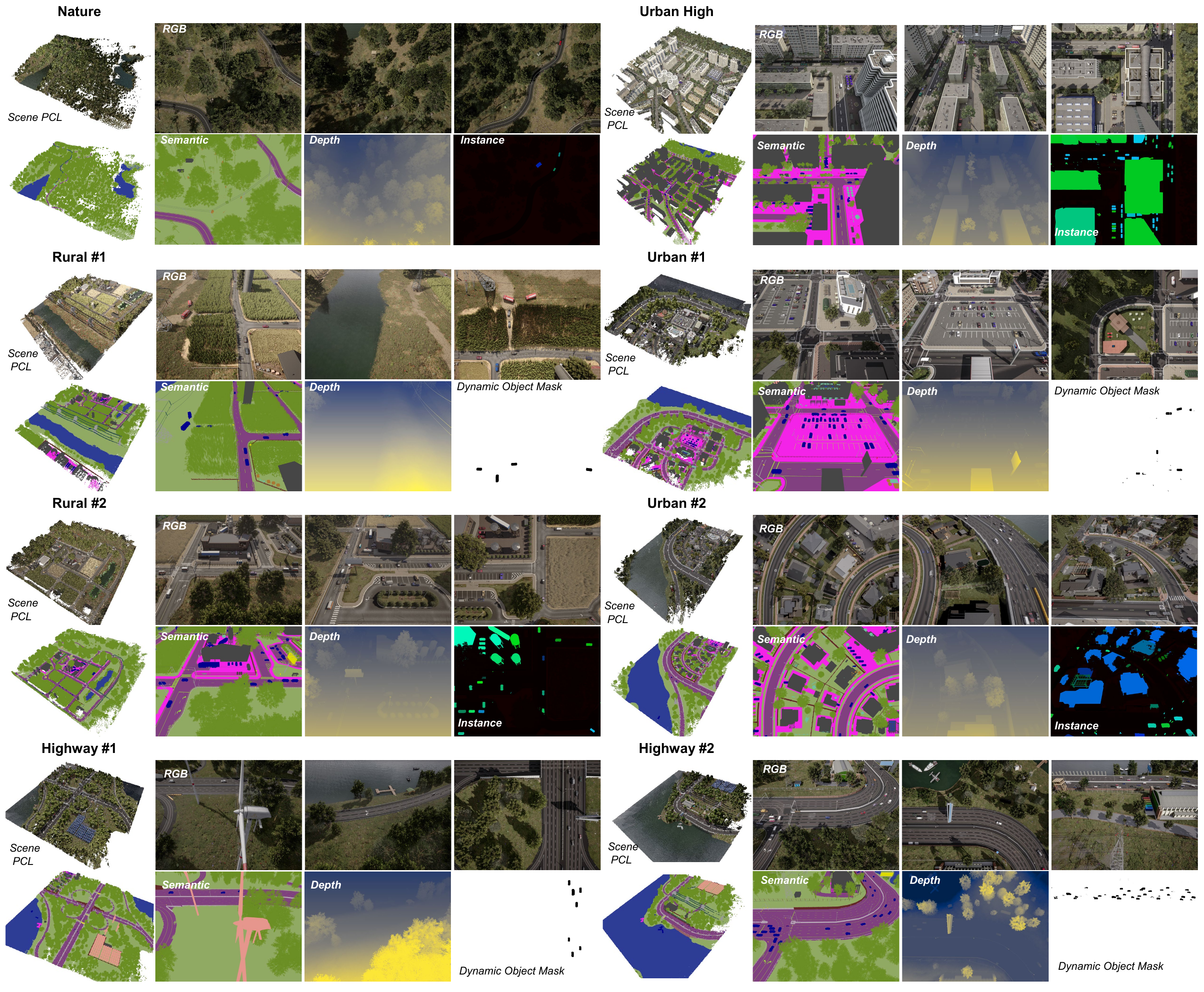}
    \caption{\textbf{ClaraVid Modalities.} Overview of the 8 UAV missions, showing representative frames with RGB, depth, semantic, instance, and dynamic annotations, along with scene-level point cloud views.}
    \label{fig:claravid_more_overview}
\end{figure*}

In this section, we present additional insights into Claravid. Additionally we present a more detailed visual overview in \autoref{fig:claravid_more_overview}.

\begin{figure}[t]
    \centering
    \includegraphics[width=\linewidth]{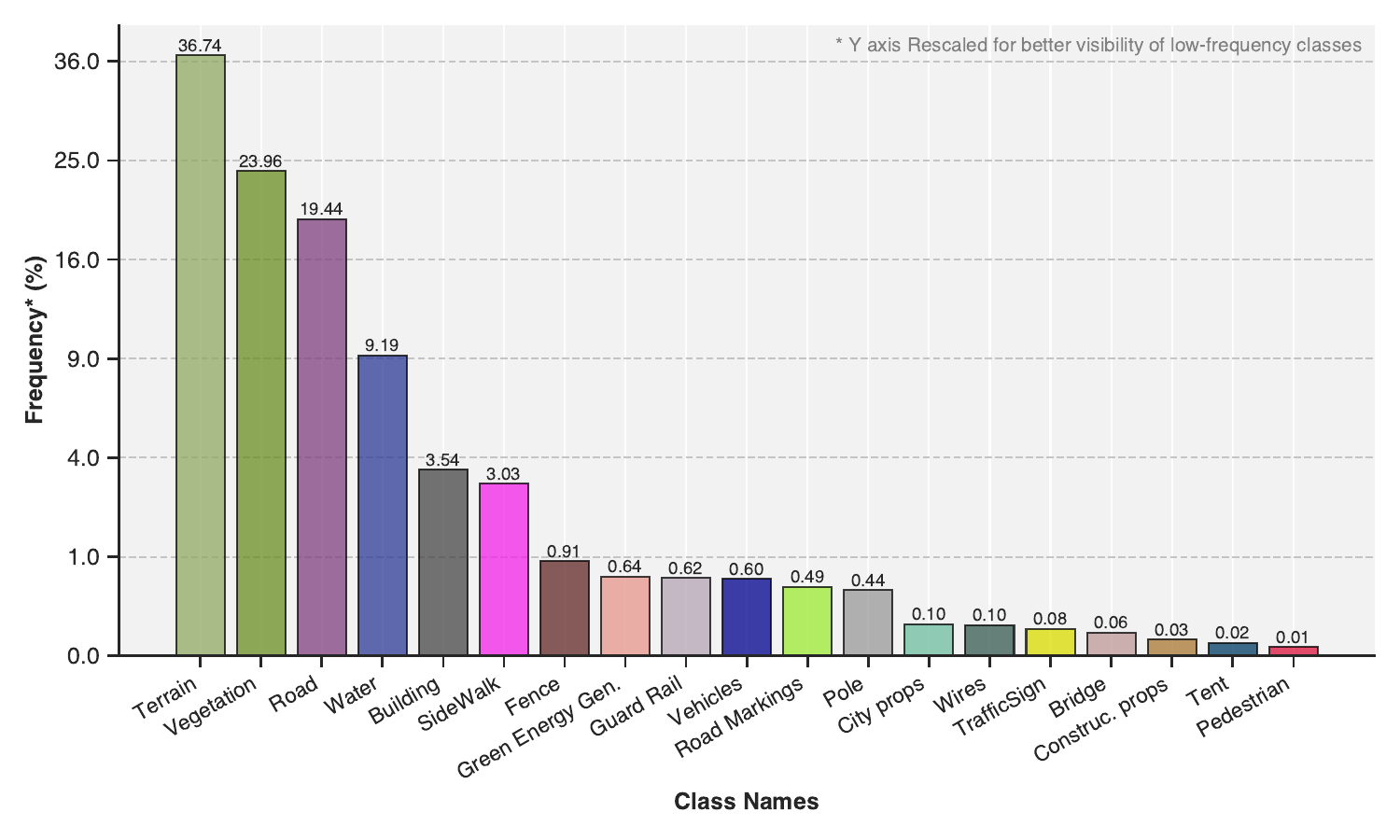}
    \caption{\textbf{Semantic Segmentation Pixel Label Distribution} The class distribution exhibits a long tail characteristic for aerial scenarios. }
    \label{fig:label_distribution}
    \vspace{-0.4cm}
\end{figure}

\subsection{Semantic Complexity}
\label{supp:scene_complexity}

\begin{figure}[t]
    \centering
    \includegraphics[width=\linewidth]{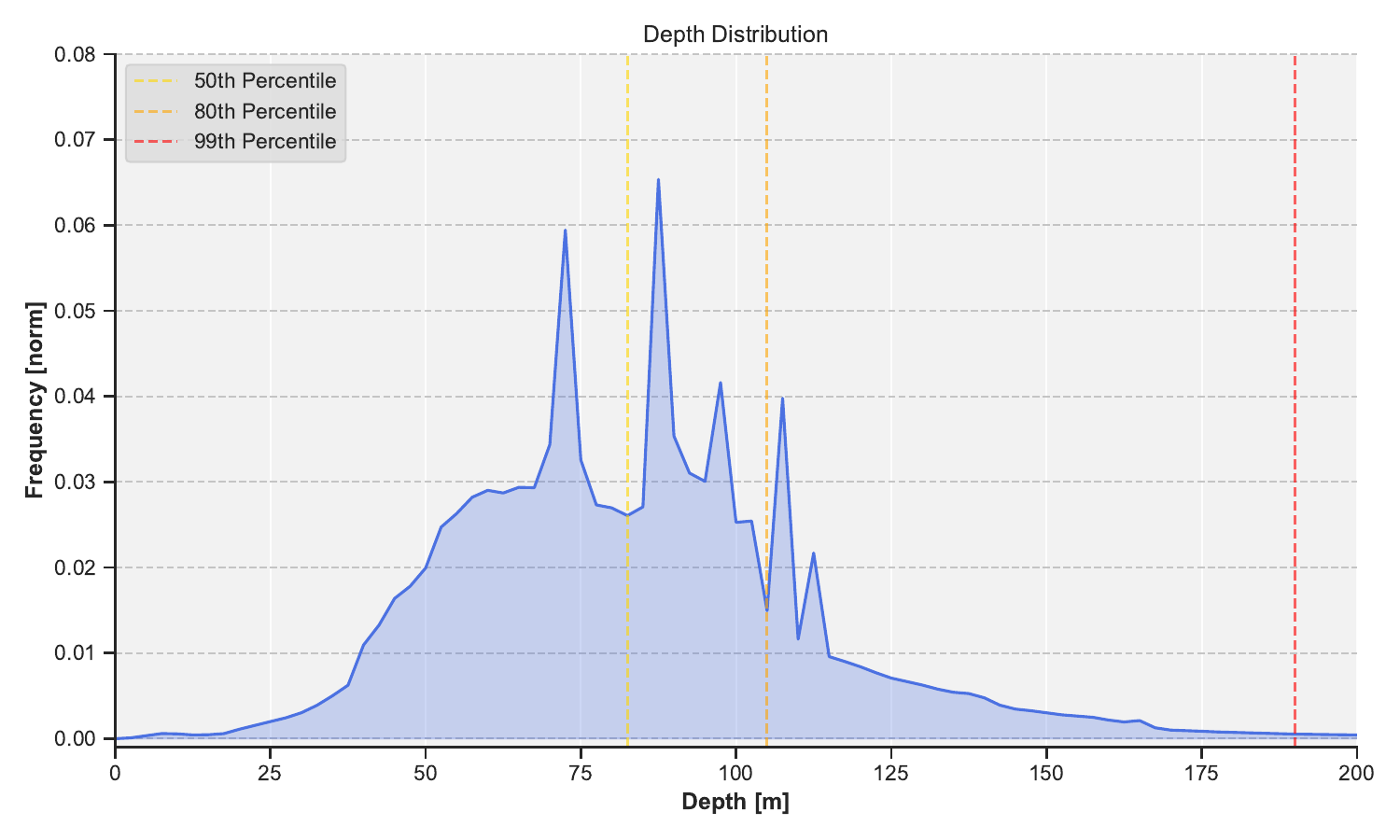}
    \caption{\textbf{Dataset Depth Distribution} The dataset’s depth values exhibit a broad distribution with prominent spikes from nadir imagery, where ground-level elements dominate. While depths range from 0–1000 m, the 99th percentile is 192[m].}
    \label{fig:depth}
    \vspace{-0.4cm}
\end{figure}

The semantic complexity of \textit{Claravid} is a direct consequence of its enriched class taxonomy, designed to capture the nuanced structures present in aerial urban and rural environments. By extending the conventional label set to include \textit{wire}, a class that encapsulates thin linear structures, and \textit{green energy}, which aggregates solar panels and renewable energy infrastructure, the dataset reflects the intricate composition of real-world landscapes. Furthermore, the \textit{pole} category has been redefined to encompass slender metallic structures such as communication towers and high-voltage power lines, thus improving its generalization for fine-grained segmentation tasks. Additionally, the introduction of \textit{urban props} and \textit{construction props} provides a finer semantic partitioning of the environment, accounting for human-centric elements in residential areas and industrial material clusters, respectively. The inclusion of a dedicated \textit{tent} class further enhances the granularity of temporary and semi-permanent structures. This expanded semantic schema not only increases the dataset's diversity but also contributes to its delentropy-based complexity profiling, allowing for a more rigorous quantification of scene heterogeneity in both structural and semantic dimensions. We present the label class distribution in \autoref{fig:label_distribution} and complementary, the depth distribution across the entire dataset in \autoref{fig:depth}.

\subsection{Rendering Pipeline Configuration}
\label{supp:rendering_pipeline_config}

To achieve high-fidelity rendering of large-scale aerial scenes in Unreal Engine 4, we implement targeted modifications to the rendering pipeline, prioritizing fine detail preservation and visibility consistency at extended distances. To mitigate the disappearance of distant objects and enhance detail clarity, we increase the rendering resolution by setting \texttt{r.ScreenPercentage=300}, effectively supersampling the scene at 3 times the native resolution. This adjustment minimizes aliasing artifacts—particularly pronounced in oblique aerial perspectives—and ensures that small-scale features remain discernible, albeit at a higher computational cost justified by the resulting visual fidelity. Shadow integrity for distant fine elements, such as foliage and thin structures, is preserved by reducing \texttt{r.Shadow.MinRadius} to 0.001, which enables shadow casting for geometrically narrow features that would otherwise be lost. Shadow map resolution is also increased to \textit{2048×2048} for objects considered having fine details, maintaining sharp shadow edges across expansive terrains. Geometric fidelity is ensured by disabling level-of-detail (LOD) transitions, preventing mesh simplification that degrades structural complexity at greater distances or altitudes, a choice prioritizing rendering quality over real-time performance scalability. Similarly, we disable distance-based object culling to maintain persistent visibility of all scene elements, eliminating sudden visibility discontinuities that disrupt spatial and temporal coherence.

% As it is closely related to the task, we also investigate the behavior of depth foundation models. for these we report the scaled results using the gt under linear transformatoin. for DepthPro we use standard resolution 1500 px, for the rest-- DepthAnything2 and Unidepth-- we use the 900 resolution infere, as going upper we obsered considerable performance downgrades.

% \begin{table*}[h]
%     \centering
%     \caption{Depth estimation performance comparison.}
%     \resizebox{\linewidth}{!}{%
%     \begin{tabular}{lcccccccc}
%         \hline
%         Depth Method & Abs Rel & Sq Rel & RMSE & RMSE Log & Log 10 err & $a_1 (<1.05)$ & $a_1 (<1.05)^2$ & $a_1 (<1.05)^3$ \\
%         \hline
%         Unidepth & 0.101 & 3.455 & 16.78 & 0.59 & 0.065 & 0.47 & 0.657 & 0.782 \\
%         Depth Anything 2 & 0.09 & 1.722 & 12.404 & 0.123 & 0.04 & 0.43 & 0.661 & 0.805 \\
%         DepthPro & 0.04 & 0.366 & 5.272 & 0.061 & 0.017 & 0.736 & 0.904 & 0.961 \\
%         \hline
%     \end{tabular}
%     }
%     \label{tab:depth_comparison}
% \end{table*}

\end{document}